\documentclass[5p]{elsarticle}
\usepackage{booktabs,shortvrb}
\usepackage{amsmath}

\usepackage{amsmath}%
\usepackage{amsfonts}%
\usepackage{amssymb}%
\usepackage{booktabs}
\usepackage{url}
\usepackage{algpseudocode}
\usepackage{algorithm}

\makeatletter
    \def\ps@pprintTitle{%
      \let\@oddhead\@empty
      \let\@evenhead\@empty
      \let\@oddfoot\@empty
      \let\@evenfoot\@oddfoot
    }
\makeatother

\newcommand{\Cluster}[0]{\ensuremath{C}}

\begin{document}

\begin{frontmatter}

\title{An Efficient Hybrid Ant Colony System for the Generalized Traveling Salesman Problem}
\author[1]{Mohammad Reihaneh}\ead{m.reihaneh@in.iut.ac.ir}
\author[2]{Daniel Karapetyan}\ead{daniel.karapetyan@gmail.com}
\address[1]{Department of Industrial Engineering, Isfahan University of Technology, Isfahan, Iran}
\address[2]{Department of Mathematics, Simon Fraser University, Surrey BC V3T\,0A3, Canada}

\begin{abstract}
The Generalized Traveling Salesman Problem (GTSP) is an extension of the well-known Traveling Salesman Problem (TSP), where the node set is partitioned into clusters, and the objective is to find the shortest cycle visiting each cluster exactly once.  In this paper, we present a new hybrid Ant Colony System (ACS) algorithm for the symmetric GTSP\@.  The proposed algorithm is a modification of a simple ACS for the TSP improved by an efficient GTSP-specific local search procedure.  Our extensive computational experiments show that the use of the local search procedure dramatically improves the performance of the ACS algorithm, making it one of the most successful GTSP metaheuristics to date.
\end{abstract}

\end{frontmatter}

\section{Introduction}
\label{sec:intro}

The Generalized Traveling Salesman Problem (GTSP) is defined as follows.  Let $V = \{ 1, 2, \ldots, n \}$ be a set of $n$ nodes being partitioned into $m$ non-empty subsets $V = C_1 \cup C_2 \cup \ldots \cup C_m$ called \emph{clusters}.  Let $\Cluster(v) = C_i$ if $v \in C_i$.  We are given a cost $d_{uv}$ of travelling between two nodes $u$ and $v$ for every $u, v \in V$ such that $\Cluster(u) \neq \Cluster(v)$.  Note that we consider only the symmetric case, i.e., $d_{uv} = d_{vu}$ for any $u, v \in V$, $\Cluster(u) \neq \Cluster(v)$.  Let $T$ be an ordered set of nodes of size $m$ such that $\Cluster(T_i) \neq \Cluster(T_j)$ for $i \neq j \in \{ 1, 2, \ldots, m \}$.  We call such a set \emph{tour}, and the weight of a tour $T$ is 
\begin{equation}
w(T) = d_{T_m, T_1} + \sum_{i = 1}^{m - 1} d_{T_i, T_{i + 1}} \,.
\end{equation}
The objective of the GTSP is to find a tour $T$ that minimizes $w(T)$.

It is sometimes convenient to consider the GTSP as a graph problem.  Let $G = (V, E)$ be a weighted undirected graph such that $(u, v) \in E$ for every $u, v \in V$ if $\Cluster(u) \neq \Cluster(v)$.  The weight of an edge $(u, v)$ is $d_{uv}$.  The objective is to find a cycle in $G$ such that it visits exactly one node in $C_i$ for $i = 1, 2, \ldots, m$ and its weight is minimized.

As a mixed integer program, the GTSP can be formulated as follows:
\begin{align*}
& \text{Minimize } \sum_{(u,v) \in E} d_{uv} \cdot x_{uv} \\
& \text{subject to } \\
& \sum_{(u,v) \in E} x_{uv} = \sum_{(u,v) \in E} x_{vu} = y_v && \text{for $v \in V$,} \\ 
& \sum_{v \in C_i} y_v = 1 && \text{for $i = 1, 2, \ldots, m$,} \\
& z_u - z_v + (m - 1) x_{uv} \le m - 2 && \text{for $(u, v) \in E$,} \\
&&& u \neq 1, v \neq 1, \\
& x_{uv} \in \{ 0, 1 \} && \text{for $(u, v) \in E$,} \\
& 1 \le z_v \le m - 1 && \text{for $v \in V \setminus \{ 1 \}$.}
\end{align*}

The GTSP is an NP-hard problem.  Indeed, if $|C_i| = 1$ for $i = 1, 2, \ldots, m$, the GTSP is reduced to the Traveling Salesman Problem (TSP)\@.  Hence, the TSP is a special case of the GTSP\@.  Since the TSP is known to be NP-hard, the GTSP is also NP-hard.

The GTSP has a lot of applications in warehouse order picking with multiple stock locations, sequencing computer files, postal routing, airport selection and routing for courier planes, and some others, see, e.g., \cite{Fischetti1997} and references therein.

Much attention was paid to the question of solving the GTSP\@.  Several researchers proposed transformations of a GTSP instance into a TSP instance, see, e.g., \cite{Ben-Arieh2003}.  At first glance, the idea of transforming a little-studied problem into a well-known one seems to be promising.  However, this approach has a limited application.  Indeed, such a transformation produces TSP instances where only the tours of some special structure correspond to feasible GTSP tours.  In particular, such tours cannot include certain edges.  This is achieved by assigning large weights to such edges making the TSP instance unusual for the exact solvers.  At the same time, solving the obtained TSP with a heuristic that does not guarantee any solution quality may produce a TSP tour corresponding to an infeasible GTSP tour.

A more efficient approach to solve the GTSP exactly is a branch-and-cut algorithm~\cite{Fischetti1997}\@.  By using this algorithm, Fischetti et al.\ solved several instances of size up to 89 clusters; solving larger instances to optimality is still too hard nowadays.  Two approximation algorithms for special cases of the GTSP were proposed in the literature; alas, the guaranteed solution quality of these algorithms is rather low for the real-world applications, see~\cite{Bontoux2010} and references therein.

In order to obtain good (but not necessarily exact) solutions for larger GTSP instances, one should consider the heuristic approach.  Several construction heuristics, discussed in~\cite{Bontoux2010,Gutin2009gtsp-memetic,Renaud1998}, generally produce low quality solutions.  A range of local searches, providing significant quality improvement over the construction heuristics, are thoroughly discussed in~\cite{Karapetyan2012gtsp-ls}.  An ejection chain algorithm exploiting the idea of the TSP Lin-Kernighan heuristic is successfully applied to the GTSP in~\cite{Karapetyan2011gtsp-lk}.  Although such complicated algorithms are able to approach the optimal solution by only several percent in less than a second for relatively large instances (the largest instance included in the test bed in~\cite{Karapetyan2011gtsp-lk} has 1084 nodes and 217 clusters), higher quality solutions may be required in practice.  In order to achieve a very high quality, one can use the metaheuristic approach.  Among the most powerful heuristics for the GTSP, there is a number of memetic algorithms, see, e.g., ~\cite{Bontoux2010,Gutin2009gtsp-memetic,Gutin2008gtsp-memetic,Silberholz2007,Snyder2006}.  Several other metaheuristic approaches were also applied to the GTSP in the literature, see, e.g., \cite{Pintea2007,Tasgetiren2007,Yang2008}.

In this paper, we focus on a metaheuristic approach called ant colony optimization (ACO)\@.  ACO was first introduced by Dorigo et al.~\cite{Dorigo1996} to solve discrete optimization problems and was inspired by the real ants behaviour.  Observe that, even without being able to see the landscape, ants are capable of finding the shortest paths between the food and the nest.  This becomes possible due to a special substance called \emph{pheromone}.  Roughly saying, an ant tends to use a path with the highest pheromone concentration.  At the beginning, there are no pheromone trails, and each ant walks randomly until it finds food.  Then it heads to the nest leaving a pheromone trail as it walks.  This pheromone trail makes this path attractive to the other ants, and so they also reach the food and walk to the nest leaving more pheromone along the path.  

An ant does not necessarily follow the pheromone trail precisely.  It may randomly select some slightly different path.  Now assume that there are several paths between the food and the nest.  The shorter is the path, the more frequent will be the walks of the ants using this path and, hence, the more pheromone it will get.  Since pheromone evaporates with time, longer paths tend to get forgotten while shorter paths tend to become popular.  Thus, in the end, most of the ants will use the shortest path.  A more detailed description of the logic staying behind the ACO algorithms can be found in~\cite{Dorigo1996} and~\cite{Dorigo2004}.

Since ants are capable of finding the shortest paths, it is natural to model their behaviour to solve such problems as the TSP or the GTSP\@.  Several metaheuristics exploiting the idea of the ant colony, are proposed in the literature.  In this study, we focus on the Ant Colony System (ACS) as it is described in~\cite{Dorigo2004}.

There are two ACO implementations for the GTSP presented in the literature.  The first one is an ACS heuristic by Pintea et al.~\cite{Pintea2007}.  It is an adaptation of the TSP ACS, and its performance is comparable with the most successful heuristics proposed by the time of its publication.  The second implementation by Yang et al.~\cite{Yang2008} is a hybrid ACS heuristic featured with a simple local search improvement procedure.

We propose a new hybrid implementation of the ACO algorithm for the GTSP\@.  The main framework of the metaheuristic is a straightforward modification of the `classical' TSP ACS implementation extended by an efficient local search procedure.  We show that such a simple heuristic is capable of reaching near-optimal solution for the GTSP instances of moderate to large sizes in a very limited time.

The paper is organized as follows.  In Section~2, we briefly present the details of the ACS algorithm for the TSP\@.  In Section~3, we propose several modifications needed to adapt the TSP algorithm for the GTSP\@.  In Section~4, we describe the local search improvement algorithm used in the metaheuristic, and in Section~5, we report and analyse the results of our computational experiments.  The outcomes of the research are summarized in Section~6.

\section{Basic ACS algorithm}
\label{sec:acs}

In this section, we briefly present the `classical' ACS algorithm as described in~\cite{Dorigo2004}.  It is described for the TSP defined by a node set $V$ of size $n$ and distances $d_{uv}$ for every pair $u \neq v \in V$.  If $w(T)$ is the weight of a Hamiltonian cycle $T$ (also called tour), the objective of the problem is to find $T$ that minimizes $w(T)$.

A hybrid ACS algorithm is a metaheuristic repeatedly constructing solutions, improving them with the local search procedure and updating the pheromone trails accordingly, see Algorithm~\ref{alg:acs}.
\begin{algorithm}[ht]
\caption{A high-level scheme of the hybrid ACS algorithm.}
\label{alg:acs}
\begin{algorithmic}
\State Initialize pheromone trails.
\While {termination condition is not met}
	\State Construct ants solutions.
	\State Apply local pheromone update.
	\State Improve the ants solutions with the local search \\
	\hskip\algorithmicindent \hskip\algorithmicindent heuristic.
	\State Save the best solution found so far.
	\State Apply global pheromone update.
\EndWhile
\end{algorithmic}
\end{algorithm}

Let $K$ be the set of ants.  The typical number of ants $|K|$ is 10.  Let $T^k$ be an ordered set of nodes corresponding to the path of the ant $k \in K$ and $T^k_i$ be the $i$th node in $T^k$.  Note that if $|T^k| = n$, the set $T^k$ can be considered as a tour.  Let $T_\text{best}$ be the best tour known so far.  Initially, we set $T_\text{best} \gets T_\text{NN}$, where $T_\text{NN}$ is the tour obtained with the Nearest Neighbor TSP heuristic, see, e.g, \cite{Gutin2008greedy} for description and discussion.

At the initialization phase, the ants are randomly distributed between the nodes: $T^k = \{ v \}$, where $v \in V$ is selected randomly for each $k \in K$.  An initial amount $\tau_0 = \frac{|K|}{w(T_\text{NN})}$ of pheromone is assigned $\tau_{uv} \gets \tau_0$ to each arc $(u, v) \in E$.  This amount has to prevent the system from a quick convergence but also should not make the convergence too slow.

On every iteration, each ant constructs a feasible TSP tour, which takes $n - 1$ steps.  Let $A^{kt} \subset V$ be the set of nodes that the ant $k \in K$ can visit on the $t$th step, $t = 1, 2, \ldots, n - 1$.  Since, in the TSP, an ant can visit any node that it did not visit before, $A^{kt} = V \setminus \{ T^k_1, T^k_2, \ldots, T^k_t \}$.  Let $\eta_{uv}$ be the so called visibility calculated as $\eta_{uv} = \frac{1}{d_{uv}}$.  Let $a_{uv} = \tau_{uv} (\eta_{uv})^\beta$, where $\beta$ is an algorithm parameter, be the value defining how much attractive is the arc $(u, v)$ for an ant.  With the probability $q_0$ (that is an algorithm parameter selected in the range $0 \le q_0 \le 1$), the ant $k$, located in the node $u = T^k_t$, selects the node $v \in A^{kt}$ that maximizes $a_{uv}$.  Otherwise it selects the node $v \in A^{kt}$ randomly, where the probability of choosing $v$ is
\begin{equation}
p^{kt}_v = \frac{a_{uv}}{\sum_{v \in A^{kt}} a_{uv}} \,.
\end{equation}

On every step of an ant $k \in K$, a local pheromone update is performed as follows:
\begin{equation}
\tau_{uv} \gets (1 - \xi) \tau_{uv} + \frac{\xi}{n \cdot w(T_\text{NN})}  \,,
\end{equation}
where $\xi$ is an algorithm parameter selected in the range $0 \le \xi \le 1$.  This update reduces the probability of visiting the arc $uv$ by the other ants, i.e., increases the chances of exploration of the other paths.

After $n - 1$ steps, each $T^k$ for $k \in K$ can be considered as a feasible TSP tour.  Run the local search improvement procedure for every $T^k$ and update the tour $T^k$ accordingly.  The typical local search improvement procedure used for the TSP is $k$-opt for $k = 2$ or $k = 3$.  Now let $k' = \arg \min_{k \in K} w(T^k)$ be the ant that performed best among $K$ in this iteration.  If $w(T_{k'}) < w(T_\text{best})$, update the best tour $T_\text{best}$ found so far with $T_{k'}$.

Finally, perform the global pheromone update.  In global pheromone update, both evaporation and pheromone deposit are applied only to the edges in the best tour $T_\text{best}$ found so far.  Let $\rho$ be an algorithm parameter called \emph{evaporation rate} and selected in the range $0 \le \rho \le 1$.  Then the global pheromone update is applied as follows:
\begin{equation}
\tau_{uv} \gets (1 - \rho) \tau_{uv} + \frac{\rho}{w(T_\text{best})} \quad \text{for $(u, v) \in T_\text{best}$\,.}
\end{equation}

Before proceeding to the next iteration, reinitialize $T^k$ with $\{ v \}$, where $v \in V$ is selected randomly for every $k \in K$.

Various termination conditions can be used in an ACS algorithm.  The most typical approaches are to limit the running time of the algorithm or to limit the number of consequent iterations in which no improvement to the original solution was found.

\section{Algorithm modifications}
\label{sec:gtsp-acs}

In order to adapt the ACS algorithm for the GTSP, we need to introduce several changes.

\begin{enumerate}
	\item The Nearest Neighbor algorithm is redefined.  Let $T^v$ for $v \in V$ be a GTSP tour obtained as follows.  Let $A$ be a set of nodes.  Set $A \gets V \setminus \Cluster(v)$.  Set $T^v \gets \{ v \}$.  On every step $t = 1, 2, \ldots, m - 1$, set $T^v_{t + 1} \gets u$ and $A \gets A \setminus \Cluster(u)$, where $u \in A$ is selected to minimize $w(T^v_t, u)$.  The output $T_\text{NN}$ of the Nearest Neighbor heuristic is the shortest tour among $T^v$, $v \in V$.
	
	\item The number $|K|$ of ants in the system is taken as an algorithm parameter and is discussed in Section~\ref{sec:results}
	
	\item Since a GTSP tour visits only $m$ nodes, the number of steps needed for an ant to construct a feasible tour is $m - 1$.
	
	\item The set $A^{kt}$ of the nodes available for the ant $k$ at the step $t$ is defined as
	$$
	A^{kt} = V \setminus \bigcup_{i = 1}^t \Cluster(T^k_i) \,.
	$$
\end{enumerate}

Let $T^i_\text{best}$ be the best tour found on or before the $i$th iteration.  The termination criteria used in our implementation is as follows: terminate the algorithm if $j \ge \Delta$ and $T^i_\text{best} = T_\text{best}$ for $i = j - \Delta + 1, j - \Delta + 2, \ldots, j$, where $j$ is the index of the current iteration and $\Delta$ is an algorithm parameter.

\section{Local Search Improvement Heuristic}
\label{sec:local-search}

It was noticed that many metaheuristics such as genetic algorithms or ant colony systems benefit from improving every candidate solution with a local search improvement procedure, see~\cite{Krasnogor2005} and references therein.  Observe that all the successful GTSP metaheuristics are, in fact, hybrid.  Thus, it is important to select an appropriate local search procedure in order to achieve a high performance.

An extensive study of the GTSP local search algorithms can be found in~\cite{Karapetyan2012gtsp-ls}.  According to the classification provided there, all the local search neighborhoods considered in the literature can be split into three classes, namely `Cluster Optimization' (CO), 'TSP-inspired' and `Fragment Optimization'.  While the latter one needs additional research in order to be applied efficiently, neighborhoods of the other two classes are widely and successfully used in the metaheuristics, see, e.g., \cite{Gutin2009gtsp-memetic,Gutin2008gtsp-memetic,Silberholz2007,Snyder2006}.

\bigskip

The CO neighborhood is the most noticeable neighborhood in the CO class.  Being of an exponential size, it can be explored in the polynomial time.  Let $T = (T_1, T_2, \ldots, T_m)$ be the given tour.  Then the CO neighborhood $N_\text{CO}(T)$ is defined as 
\begin{multline}
N_\text{CO}(T) = \big\{ (T'_1, T'_2, \ldots, T'_m) :\ \\ T'_i \in \Cluster(T_i) \text{ for } i = 1, 2, \ldots, m \big\} \,.
\end{multline}
Note that the size of the CO neighborhood is 
$$
|N_\text{CO}(T)| = \prod_{i = 1}^m |C_i| \in O(s^m) \,,
$$
where $s = \max_{i = 1}^m |C_i|$ is the size of the largest cluster in the problem instance.  Next we will briefly explain the CO algorithm finding the shortest tour $T' \in N_\text{CO}(T)$.

Let $T = (T_1, T_2, \ldots, T_m)$ be the given tour.  Create a copy $S$ of the cluster $\Cluster(T_1)$.  Construct a multilayer directed graph $G_\text{CO}(V_\text{CO}, E_\text{CO})$ with the layers $\Cluster(T_1)$, $\Cluster(T_2)$, 
\ldots, $\Cluster(T_m)$, $S$.  For every pair of consecutive layers $L_1$ and $L_2$, for every pair of vertices $u \in L_1$ and $v \in L_2$, create an arc $(u, v)$ of weight $d_{uv}$.  Let $P_v$ be the shortest path from $v \in \Cluster(T_1)$ to its copy $v' \in S$.  Note that $P_v$ corresponds to a tour visiting the clusters in the same order as $T$ does.  Select $v \in \Cluster(T_1)$ that minimizes the weight of $P_v$.  The corresponding cycle is the shortest tour $T' \in N_\text{CO}(T)$, and the procedure terminates in $O(n s^2)$ time.

Several heuristic improvements of the above algorithm were proposed~\cite{Karapetyan2012gtsp-ls}.  In this research, we implemented only the easiest and the most important one.  Note that the complexity of the algorithm linearly depends on the size of the cluster $\Cluster(T_1)$.  Since a tour can be arbitrarily rotated, let $\Cluster(T_1)$ be the smallest cluster.  This modification reduces the time complexity of the CO algorithm to $O(n \gamma s)$, where $\gamma = \min_{i = 1}^m |C_i|$ is the size of the smallest cluster.

\bigskip

Recall that the most typical neighborhoods used for the TSP are $k$-opt.  Several adaptation of the TSP $k$-opt were proposed in~\cite{Karapetyan2012gtsp-ls}, and the resulting neighborhoods were classified as `TSP-inspired'.  Since we aim at designing a fast and simple metaheuristic, we chose the `Basic' $k$-opt adaptation~\cite{Karapetyan2012gtsp-ls}.  In short, let $\mathcal{P}_\text{GTSP}$ be the original GTSP and let $T = (T_1, T_2, \ldots, T_m)$ be the given tour defined in $\mathcal{P}_\text{GTSP}$.  Let $G_\text{TSP}(V_\text{TSP}, E_\text{TSP})$ be the complete subgraph of $G$, where $V_\text{TSP} = \{ T_1, T_2, \ldots, T_m \}$.  Construct a TSP $\mathcal{P}_\text{TSP}$ for the graph $G_\text{TSP}$.  Note that the tour $T$ defined for $\mathcal{P}_\text{GTSP}$ is a feasible tour of the same weight in $\mathcal{P}_\text{TSP}$, and any feasible tour in $\mathcal{P}_\text{TSP}$ is a feasible tour of the same weight in $\mathcal{P}_\text{GTSP}$\@.  Improve the tour $T$ with the TSP $k$-opt algorithm.  The obtained tour is the result of the `Basic' adaptation of the $k$-opt local search.

\bigskip

It was shown that a combination of neighborhoods of different classes is often superior to the component local searches~\cite{Karapetyan2011map-ls}.  Thus, we use a local search that combines the neighborhoods of the CO and the 'TSP-inspired' classes.  In particular, the improvement procedure used in our algorithm proceeds as follows.  First, the given tour is improved with the `Basic' adaptation of the 3-opt local search.  Then, the CO algorithm is applied to it.  No further optimization is performed so that the resulting solution is not guaranteed to be a local minimum with regards to the 3-opt neighborhood.  

This local search procedure was obtained empirically after extensive computational experiments with different local search neighborhoods and strategies.

\section{Computational Experiments}
\label{sec:results}

As a part of our research, we conducted extensive computational experiments to find the best parameter values and to measure the algorithm's performance.  Our testbed includes a number of instances produced from the standard TSP benchmark instances by applying a simple clustering procedure proposed in~\cite{Fischetti1997}.  Such an approach was used by many researchers, see, e.g., \cite{Bontoux2010,Gutin2009gtsp-memetic,Karapetyan2012gtsp-ls}.  We selected the same set of instances as in~\cite{Bontoux2010} and \cite{Silberholz2007}.  Our ACS algorithm and the local search procedures are implemented in C\# and the computational platform is based on 2.93~GHz Intel Core 2 Due CPU\@.

We used the following values of the algorithm parameters: $\beta = 3$, $\rho = 0.4$, $\xi = 0.03$, $q_0 = 0$, $\Delta = 300$ and $|K| = 10$.  Among all the combinations of $\beta$, $\rho$, $\xi$, $q_0$, $\Delta$ and $|K|$ that we tried, this one provided the best, on average, experimental results.  However, we noticed that slight variations of these values do not significantly change the behaviour of the metaheuristic.

The extension of the local search procedure with the CO algorithm is the most significant modification implemented in our ACS\@.  Thus, we start from studying the impact of the CO algorithm on the performance of the ACS\@.  In our first series of experiments, we show the importance of this modification.  In what follows, HACS refers to our hybrid ACS metaheuristic with the composite local search procedure as described above, and HACS$_0$ refers to the simplified version of the metaheuristic that uses only the 3-opt algorithm as the local search procedure.

The HACS and the HACS$_0$ algorithms are compared in Table~\ref{tab:co}.
\begin{table*}[ht]
\begin{center}
\begin{tabular}{@{} lr @{} c @{} rr @{} c @{} rr @{} c @{} rr @{}}
\toprule
&& \hspace{2em} & \multicolumn{2}{@{} c @{}}{Error, \%} & \hspace{2em} & \multicolumn{2}{@{} c @{}}{Time, sec} & \hspace{2em} & \multicolumn{2}{@{} c @{}}{Optimal, \%} \\
\cmidrule(){4-5}
\cmidrule(){7-8}
\cmidrule(){10-11}
Instance & Best && HACS & HACS$_0$ && HACS & HACS$_0$ && HACS & HACS$_0$ \\
\midrule
40d198	&	10557	&&	\underline{0.00}	&	0.69	&&	\underline{2.74}	&	7.45	&&	\underline{100}	&	0	\\
40kroa200	&	13406	&&	\underline{0.00}	&	0.62	&&	\underline{2.43}	&	6.57	&&	\underline{90}	&	10	\\
40krob200	&	13111	&&	\underline{0.00}	&	1.22	&&	\underline{2.55}	&	5.05	&&	\underline{90}	&	0	\\
45ts225	&	68340	&&	\underline{0.01}	&	0.73	&&	\underline{2.71}	&	5.87	&&	\underline{40}	&	10	\\
46pr226	&	64007	&&	\underline{0.00}	&	0.06	&&	\underline{2.29}	&	4.69	&&	\underline{100}	&	20	\\
53gil262	&	1013	&&	\underline{0.41}	&	1.99	&&	\underline{5.57}	&	9.38	&&	\underline{60}	&	0	\\
53pr264	&	29549	&&	\underline{0.00}	&	0.83	&&	\underline{3.83}	&	12.89	&&	\underline{100}	&	0	\\
60pr299	&	22615	&&	\underline{0.03}	&	0.58	&&	\underline{5.98}	&	11.98	&&	\underline{60}	&	0	\\
64lin318	&	20765	&&	\underline{0.00}	&	2.37	&&	\underline{4.87}	&	15.95	&&	\underline{100}	&	10	\\
80rd400	&	6361	&&	\underline{0.62}	&	3.90	&&	\underline{9.95}	&	32.05	&&	\underline{20}	&	0	\\
84fl417	&	9651	&&	\underline{0.00}	&	0.11	&&	\underline{7.22}	&	31.35	&&	\underline{100}	&	0	\\
88pr439	&	60099	&&	\underline{0.00}	&	0.87	&&	\underline{10.06}	&	40.24	&&	\underline{100}	&	0	\\
89pcb442	&	21657	&&	\underline{0.09}	&	2.25	&&	\underline{13.41}	&	38.51	&&	\underline{30}	&	0	\\
99d493	&	20023	&&	\underline{0.51}	&	2.04	&&	\underline{22.68}	&	53.91	&&	\underline{0}	&	\underline{0}	\\
107att532	&	13464	&&	\underline{0.15}	&	1.04	&&	\underline{17.82}	&	58.95	&&	\underline{20}	&	0	\\
107si535	&	13502	&&	\underline{0.02}	&	1.02	&&	\underline{19.99}	&	67.60	&&	\underline{60}	&	0	\\
113pa561	&	1038	&&	\underline{0.13}	&	2.94	&&	\underline{19.26}	&	54.71	&&	\underline{10}	&	0	\\
115rat575	&	2388	&&	\underline{1.52}	&	4.13	&&	\underline{26.79}	&	74.04	&&	\underline{10}	&	0	\\
131p654	&	27428	&&	\underline{0.00}	&	0.11	&&	\underline{18.57}	&	90.30	&&	\underline{100}	&	0	\\
132d657	&	22498	&&	\underline{0.21}	&	2.90	&&	\underline{37.43}	&	138.85	&&	\underline{0}	&	\underline{0}	\\
145u724	&	17272	&&	\underline{1.57}	&	4.14	&&	\underline{48.80}	&	137.71	&&	\underline{0}	&	\underline{0}	\\
157rat783	&	3262	&&	\underline{1.37}	&	4.99	&&	\underline{47.41}	&	181.94	&&	\underline{0}	&	\underline{0}	\\
201pr1002	&	114311	&&	\underline{0.28}	&	2.46	&&	\underline{123.38}	&	364.24	&&	\underline{10}	&	0	\\
207si1032	&	22306	&&	\underline{0.37}	&	4.44	&&	\underline{177.00}	&	305.92	&&	\underline{0}	&	\underline{0}	\\
212u1060	&	106007	&&	\underline{0.66}	&	2.38	&&	\underline{103.89}	&	371.33	&&	\underline{0}	&	\underline{0}	\\
217vm1084	&	130704	&&	\underline{0.66}	&	2.46	&&	\underline{95.35}	&	409.04	&&	\underline{20}	&	0	\\
\midrule
Average	&		&&	\underline{0.33}	&	1.97	&&	\underline{32.00}	&	97.33	&&	\underline{47}	&	2	\\
\bottomrule
\end{tabular}
\end{center}
\setcounter{table}{0}
\caption{Comparison of the HACS algorithm with its simplified version HACS$_0$.}
\label{tab:co}
\end{table*}
The columns of the table are as follows:
\begin{enumerate}
	\item `Instance' is the name of the the GTSP test instance.  It consists of three parts, namely the number of clusters $m$, the type of the instance (derived from the original TSP instance) and the number of vertices $n$.  
	\item `Best' is the objective of the best solution known so far for the given problem instance.  For the instances of size $m \le 89$ the optimal solutions are known, see~\cite{Fischetti1997}.  For the other instances the values are taken from~\cite{Gutin2009gtsp-memetic}.
	\item `Error' is the relative solution error $e$, in percent, calculated as follows: 
	$$
	e = \frac{w(T) - w(T_\text{best})}{w(T_\text{best})} \cdot 100\% \,,
	$$
	where $T$ is the solution to be evaluated and $T_\text{best}$ is the best solution known so far.
	\item `Time' is the running time of the algorithm.
	\item `Optimal' is the number of runs, in percent, in which the best known so far solution was obtained.
\end{enumerate}
The best result in a row is underlined.  Since the ACO algorithms are non-deterministic, in order to get some statistically significant results we repeat every experiment 10 times.  Hence, every result reported in Table~\ref{tab:co} is an average over the 10 runs.

It is easy to see that the full version of the HACS clearly dominates the simplified one.  This shows the importance of selecting the optimal nodes within clusters and also proves the efficiency of the approach used in our local search improvement procedure.  It is worth noting that a more common adaptation of a TSP local search for the GTSP is to hybridize the `TSP-inspired' and `Cluster Optimization' neighborhoods~\cite{Karapetyan2012gtsp-ls,Renaud1998}.  However, our experiments prove that applying two local searches of different classes one after another may be a more effective strategy.

\bigskip

In order to evaluate the efficiency of the HACS, we compare its performance to the performance of several other metaheuristics, see Table~\ref{tab:compare}.
\setcounter{table}{1}
\begin{table*}[ht]
\begin{center}
\begin{tabular}{@{} l @{} c @{} rrrr @{} c @{} rrr @{}}
\toprule
& \hspace{1em} & \multicolumn{4}{@{} c @{}}{Error, \%} & \hspace{2em} & \multicolumn{3}{@{} c @{}}{Normalized time, sec} \\
\cmidrule(){3-6}
\cmidrule(){8-10}
Instance	&&	HACS	&	SG	&	BAF	&	PPC	&&	HACS	&	SG	&	BAF	\\
\midrule
40d198	&&	\underline{0.00}	&	\underline{0.00}	&	\underline{0.00}	&	0.01	&&	2.74	&	\underline{1.09}	&	10.15	\\
40kroa200	&&	\underline{0.00}	&	\underline{0.00}	&	\underline{0.00}	&	0.01	&&	2.43	&	\underline{1.11}	&	10.41	\\
40krob200	&&	\underline{0.00}	&	0.05	&	\underline{0.00}	&	\underline{0.00}	&&	2.55	&	\underline{1.09}	&	10.81	\\
45ts225	&&	\underline{0.01}	&	0.14	&	0.04	&	0.03	&&	2.71	&	\underline{1.14}	&	31.45	\\
46pr226	&&	\underline{0.00}	&	\underline{0.00}	&	\underline{0.00}	&	0.03	&&	2.29	&	\underline{1.03}	&	8.25	\\
53gil262	&&	0.41	&	0.45	&	\underline{0.14}	&	0.22	&&	5.57	&	\underline{2.43}	&	24.34	\\
53pr264	&&	\underline{0.00}	&	\underline{0.00}	&	\underline{0.00}	&	0.00	&&	3.83	&	\underline{1.57}	&	18.27	\\
60pr299	&&	0.03	&	0.05	&	\underline{0.00}	&	0.24	&&	5.98	&	\underline{3.06}	&	21.25	\\
64lin318	&&	\underline{0.00}	&	\underline{0.00}	&	\underline{0.00}	&	0.12	&&	\underline{4.87}	&	5.39	&	26.33	\\
80rd400	&&	0.62	&	0.58	&	\underline{0.42}	&	0.87	&&	9.95	&	\underline{9.72}	&	32.21	\\
84fl417	&&	\underline{0.00}	&	0.04	&	\underline{0.00}	&	0.57	&&	7.22	&	\underline{5.43}	&	31.63	\\
88pr439	&&	\underline{0.00}	&	\underline{0.00}	&	\underline{0.00}	&	0.78	&&	\underline{10.06}	&	12.71	&	42.55	\\
89pcb442	&&	0.09	&	\underline{0.01}	&	0.19	&	0.69	&&	\underline{13.41}	&	15.62	&	62.53	\\
99d493	&&	0.51	&	0.47	&	\underline{0.44}	&	---	&&	\underline{22.68}	&	23.81	&	166.10	\\
107att532	&&	0.15	&	0.35	&	\underline{0.05}	&	---	&&	\underline{17.82}	&	21.13	&	137.54	\\
107si535	&&	\underline{0.02}	&	0.08	&	0.07	&	---	&&	19.99	&	\underline{17.57}	&	90.98	\\
113pa561	&&	\underline{0.13}	&	1.50	&	0.42	&	---	&&	19.26	&	\underline{14.05}	&	149.43	\\
115rat575	&&	1.52	&	\underline{1.12}	&	1.16	&	---	&&	\underline{26.79}	&	32.32	&	157.01	\\
131p654	&&	\underline{0.00}	&	0.29	&	0.01	&	---	&&	\underline{18.57}	&	21.78	&	144.95	\\
132d657	&&	\underline{0.21}	&	0.45	&	0.30	&	---	&&	\underline{37.43}	&	88.16	&	259.11	\\
145u724	&&	1.57	&	\underline{0.57}	&	1.02	&	---	&&	\underline{48.80}	&	107.88	&	218.66	\\
157rat783	&&	1.37	&	1.17	&	\underline{1.10}	&	---	&&	\underline{47.41}	&	101.43	&	391.79	\\
201pr1002	&&	0.28	&	\underline{0.24}	&	0.27	&	---	&&	\underline{123.38}	&	309.57	&	513.48	\\
207si1032	&&	0.37	&	0.37	&	\underline{0.11}	&	---	&&	177.00	&	\underline{161.58}	&	616.28	\\
212u1060	&&	\underline{0.66}	&	2.25	&	1.31	&	---	&&	\underline{103.89}	&	396.43	&	762.86	\\
217vm1084	&&	0.66	&	0.90	&	\underline{0.64}	&	---	&&	\underline{95.35}	&	374.69	&	583.44	\\
\midrule
Average (all)	&&	0.33	&	0.43	&	\underline{0.30}	&	---	&&	\underline{32.00}	&	66.61	&	173.92	\\
Average ($m \le 89$)	&&	0.09	&	0.10	&	\underline{0.06}	&	0.27	&&	5.66	&	\underline{4.72}	&	25.40	\\
\bottomrule
\end{tabular}
\end{center}
\setcounter{table}{1}
\caption{Comparison of the HACS algorithm with the other GTSP metaheuristics.}
\label{tab:compare}
\end{table*}
In particular, we compare the HACS to three other metaheuristics, namely the memetic algorithm SG by Silberholz and Golden~\cite{Silberholz2007}, a memetic algorithm BAF by Bontoux et el.~\cite{Bontoux2010} and an ACO algorithm PPC by Pintea et el.~\cite{Pintea2007}.  

The running times of SG and BAF reported in Table~\ref{tab:compare} are normalized to compensate the difference in the experimental platforms.  The SG algorithm was implemented in Java and tested on a machine with 3~GHz Intel Pentium 4 CPU which we estimate to be approximately 1.5 times slower than our platform.  The BAF algorithm was implemented in C++ and tested on a machine with 2~GHz Intel Pentium 4 CPU which we estimate to be similar to our platform (note that the C++ implementations are often considered to be twice faster than the Java or C\# implementations~\cite{Gutin2009gtsp-memetic}).  The running time of PPC for each of the instances is 10~minutes as this was the termination criteria chosen in~\cite{Pintea2007} (the computational platform is not reported in~\cite{Pintea2007}).

For all the SG, BAF and PPC algorithms, the reported values are the averages among 5 runs; the results of HACS are the averages among 10 runs.

Since the results of the PPC algorithm are reported for only a subset of the instances in our testbed, we provide two averages in every column of Table~\ref{tab:compare}.  The first average (denoted as `all') is the average over all the instances in our testbed, i.e., $40 \le m \le 217$.  The second average (denoted as $m \le 89$) is the average over the testbed chosen in~\cite{Pintea2007}, i.e., $40 \le m \le 89$.

\bigskip

In fact, we also compared our HACS to the ACO algorithm YSML by Yang et el.~\cite{Yang2008} and a memetic algorithm GK by Gutin and Karapetyan~\cite{Gutin2009gtsp-memetic}, though those results are excluded from Table~\ref{tab:compare}.  

The results reported in~\cite{Yang2008} are obtained for the instances of size $10 \le m \le 40$ (the testbed was generated from the TSP instances by using the same clustering procedure).  It was noticed that these instances are relatively easy to solve to optimality even with a local search procedure, see~\cite{Karapetyan2011gtsp-lk}.  Our ACS also solves all these instance to optimality and takes at most 1~sec for each run.  The running time of YSML is not reported in~\cite{Yang2008}, but the solutions obtained in~\cite{Yang2008} are often not optimal.  We conclude that our algorithm outperforms YSML.

The GK memetic algorithm~\cite{Gutin2009gtsp-memetic} is the state-of-the-art algorithm that, until now, was not outperformed by any other metaheuristic.  It is a sophisticated heuristic with a well-tuned local search improvement procedure and innovative genetic operators.  Although GK dominates the HACS with respect to both the solution quality and the running time, it does not affect the outcomes of our research.  Indeed, we aim at showing that a simple modification of the `classical' ACO algorithm can yield an efficient solver for a hard combinatorial optimization problem.  Also note that HACS and GK belong to the different classes of metaheuristics.

\bigskip

Table~\ref{tab:compare} shows that our HACS algorithm is similar to SG and BAF and significantly outperforms PPC with regards to the solution quality.  Although, on average, BAF performs slightly better than HACS, there is no clear domination since for some instances the HACS produces better solutions than BAF does.  Similarly, SG is dominated by neither HACS nor BAF\@.  With regards to the running time, HACS is the fastest heuristic for the large instances while SG usually takes less time for the instances of size $m \le 84$.  The BAF algorithm is the slowest one in every experiment and, on average, it is 5 times slower than HACS\@.  

Note that the above comparison of the running times is rather inaccurate since the considered algorithms were tested on different platforms, and only a rough normalization of the running times was performed.  Still, certain outcomes can be made.  In particular, the SG algorithm performs very well for the small instances while it is outperformed by HACS for larger instances with regards to both the solution quality and the running time.  BAF, on average, produces better solutions then either HACS or SG do but this is achieved at the cost of significantly larger running times.  Finally, HACS is superior to the other ACO algorithms, namely PPC and YSML, though the comparison was only possible for a limited number of test instances.

\section{Conclusions}
\label{sec:conclusions}

An efficient ACO heuristic for the GTSP is proposed in this paper.  It is obtained from a `classical' TSP ACS algorithm by several straightforward modifications and hybridisation with a simple local search procedure.  It was shown that, among other reasons, the success of our HACS is due to the effective combination of two local search heuristics of different classes.  Extensive computational experiments were conducted in order to prove that HACS performs as well as the most successful memetic algorithms proposed for the GTSP with the exception of the state-of-the-art sophisticated metaheuristic.  It was also shown that HACS outperforms two other ACO GTSP algorithms proposed in the literature.

\section*{Acknowledgement}

We would like to thank Prof.~Mohammad S.~Sabbagh for his very helpful comments and suggestions.


\bibliographystyle{abbrv}
\bibliography{../../gtsp,../../map,../../metaheuristics}{}

\end{document}